# DYNAMICALLY HANDLING TASK DISRUPTIONS BY COMPOSING TOGETHER BEHAVIOR MODULES

Thomas E. Portegys, portegys@gmail.com , ORCID 0000-0003-0087-6363

Dialectek, DeKalb, Illinois USA

## ABSTRACT

Biological neural networks operate in the presence of task disruptions as they guide organisms toward goals. A familiar stream of stimulus-response causations can be disrupted by subtask streams imposed by the environment. For example, taking a familiar path to a foraging area might be disrupted by the presence of a predator, necessitating a "detour" to the area. The detour can be a known alternative path that must be dynamically composed with the original path to accomplish the overall task. In this project, overarching base paths are disrupted by independently learned path modules in the form of insertion, substitution, and deletion modifications to the base paths such that the resulting modified paths are novel to the network. The network's performance is then tested on these paths that have been learned in piecemeal fashion. In sum, the network must compose a new task on the fly. Several network architectures are tested: Time delay neural network (TDNN), Long short-term memory (LSTM), Temporal convolutional network (TCN), and Morphognosis, a hierarchical neural network. LSTM and Morphognosis perform significantly better for this task.

## INTRODUCTION

Life is predictable until it isn't. Behaving flexibly is obviously an advantage in dealing with environments that change. Learning is one form of flexibility (Dridi and Lehmann, 2016). Another form, explored in this project, is how modules of learned behavior can be combined at "run time" to accomplish tasks, a skill that humans and many animals are adept at (Clune et al., 2013; Lorenz et al., 2011).

Biological neural networks operate in the presence of task disruptions as they guide animals toward goals in dynamically changing environments. For example, taking a familiar path to a foraging area might be disrupted by the presence of a predator, necessitating a "detour" to the area. The detour can be a known alternative path that must be dynamically composed with the original path to accomplish the overall task.

Animal task performance in changing environments is largely about learning. For example, it has been found that rodents learn mazes that are similar to previously learned mazes more quickly (Rama, et al., 2016; Alonso et al., 2021). Memories can be patched together with new learning to accelerate the process. Rosenberg et al. (2021) found that mice are able to quickly solve a labyrinth task before having a global memory formed by piecing together local turning rules. Other studies have found that innate behavioral "modules" can be dynamically assembled into hunting behavior in zebrafish (Mearns et al., 2019) and foraging in mice (Hörndli et al., 2018).

In this project, the performance of several types of artificial neural networks (ANNs) are tested as trained overarching base paths are disrupted by independently trained path modules in the form of insertion, substitution, and deletion modifications to the base paths. The resulting modified paths are novel to the networks. A module is defined as a sequence of stimulus-response causations that allow an agent to navigate from a start to an end state.

Modularity is a long-standing goal for machine learning (Baldwin & Clark, 2000; Csordás et al., 2020; Melin et al., 2005; Rojas et al., 1996) as it allows for faster learning and hierarchically composing behavior. In contrast, modular processing is an innate ability in humans. For example, children will very quickly learn to identify a novel animal by piecing together parts of similar animals that they know such as ears, eyes, etc. ANNs typically require a full relearning to solve such tasks.

The author has an abiding interest in ANN modularity. Previous projects have compared the performance of various architectures on a goal-seeking maze tasks in which modular segments that are learned independently must be composed to achieve a goal while retaining long-term state information essential to finalizing the task (Portegy, 2010; Portegys, 2021). The current project is an examination of the resiliency of long-term task memory that is disrupted by subtasks that force "detours". Various ANNs architectures are tested, including Time delay neural network (TDNN), Temporal convolutional neural network (TCN), Long short-term memory (LSTM), and Morphognosis, a hierarchical neural network.

## DESCRIPTION

The system is composed of two components: a path composer that generates paths and a learning component, consisting of TDNN, TCN, LSTM and Morphognosis neural networks. The code is available at https://github.com/morphognosis/TaskComposer.

### PATH COMPOSER

The path composer component generates datasets for training and testing by the ANNs. A dataset consists of set of paths. Each path is a sequence of stimuli consisting of random numbers concatenated with path number and corresponding response numbers. The agent learns to produce the correct response at each step in the sequence.

Training data:

- A base path. This path represents the overall task to be performed.
- A variable number of modular paths.

Testing data:

- A variable number of paths created by inserting, substituting, and deleting portions of the base path with randomly chosen modular paths. A single disruption is performed for each test path:
    - Insertion: a modular path is injected at a random position in the base path and the base path after the modular path is shifted right.
    - Substitution: a modular path replaces a segment of the base path at a random position.
    - Deletion: A segment having the length of a modular path is deleted from the base path at a random position and the base path after the modular path is shifted left.

A sample dataset is shown in Figure 1.

```
Training paths:
Base path:
stimuli: 0:8 0:9 0:6 0:7 0:7 0:0 0:9 0:10 0:8 0:11 0:8 0:9 0:4 0:14 0:14
responses: 3 10 6 10 4 6 13 10 3 9 10 2 9 2 8
Modular paths:
stimuli: 1:12 1:12
responses: 2 11
stimuli: 2:14 2:7 2:7
responses: 3 10 4
stimuli: 3:2 3:9 3:7 3:8
responses: 11 4 11 14
stimuli: 4:3 4:11 4:2
responses: 11 3 1
stimuli: 5:7 5:6 5:7
```

```
responses: 7 12 14

Test paths:

Insertion:

stimuli: 0:8 0:9 0:6 0:7 0:7 0:0 0:9 0:10 0:8 0:11 0:8 0:9 0:4 0:14 0:14 2:14 2:7 2:7

responses: 3 10 6 10 4 6 13 10 3 9 10 2 9 2 8 3 10 4

stimuli: 0:8 0:9 0:6 0:7 5:7 5:6 5:7 0:7 0:0 0:9 0:10 0:8 0:11 0:8 0:9 0:4 0:14 0:14

responses: 3 10 6 10 7 12 14 4 6 13 10 3 9 10 2 9 2 8

Substitution:

stimuli: 0:8 0:9 0:6 0:7 0:7 0:0 0:9 0:10 0:8 0:11 0:8 2:14 2:7 2:7 0:14

responses: 3 10 6 10 4 6 13 10 3 9 10 3 10 4 8

stimuli: 0:8 0:9 0:6 0:7 0:7 0:0 0:9 0:10 0:8 0:11 0:8 0:9 5:7 5:6 5:7

responses: 3 10 6 10 4 6 13 10 3 9 10 2 7 12 14

Deletion:

stimuli: 0:7 0:7 0:0 0:9 0:10 0:8 0:11 0:8 0:9 0:4 0:14 0:14

responses: 10 4 6 13 10 3 9 10 2 9 2 8

stimuli: 0:8 0:9 0:6 0:10 0:8 0:11 0:8 0:9 0:4 0:14 0:14

responses: 3 10 6 10 3 9 10 2 9 2 8
```

Figure 1 – Sample dataset.

Stimuli and responses are one-hot encodings.

The LSTM and TCN datasets have this shape:

[<number of paths>, <path length>, <stimulus/response size>]

TDNN uses a sliding window such that the network sees a sequence as a subsequence of paths shifting in one input at a time, with the entire path presented at each step. Thus the input shape:

[<number of paths> * <path length>, <path length> * <stimulus size>]

and the output shape:

[<number of paths> * <path length>, <response size>]

Data for the Morphognosis network, described in the next section, is organized as for TDNN, but is also preprocessed by aggregating, skewing and normalizing time intervals. A time interval is a container for multiple stimuli that can span one or more steps in a sequence.

For intervals that span a single step, stimuli are aggregated as follows:

$$interval_t = \{ stimulus_t, stimulus_{t-1}, \ldots, stimulus_0 \} \qquad (1)$$

Where $t$ is the number of time steps prior to the current ($t$ = 0) step. Thus the current step contains a single stimulus input to the network.

The skewing operation allows an interval to span multiple time steps such multiple stimuli can be mapped to the same interval. The process begins by assigning a weight of 1 to each interval. Then the weights are shifted back in time for each interval, starting with $interval_t$:

$$w = (\sum_0^{t-1} weight_i) * SKEW \qquad (2)$$

$$weight_t = weight_t + w \qquad (3)$$

$$weight_{i<t} = weight_{i<t} - \left(\frac{w}{t}\right) \qquad (4)$$

Where *SKEW* is a constant between 0 (no skew) and 1 (maximum skew).

As an example, a *SKEW* of .5 and 13 time steps results in the following weights:

```
[7.5 3.5 1.625 0.75 0.34375 0.15625 0.0703125 0.03125 0.013671875
0.005859375 0.0024414062 9.765625E-4 3.6621094E-4]
```

The integer value of an interval's weight then represents the number of time steps mapped to that interval. Intervals with no mappings are eliminated. This results in the following 5 intervals:

*[t, t, t, t, t, t, t], [t-1, t-1, t-1], [t-2], [t-3], [t-4]*

Thus time steps 6 through 12 are mapped to the same interval containing the oldest stimuli, steps 5 through 3 to the next most recent interval, etc.

The stimuli values are added into their respective intervals, and then the values in the interval are normalized to a maximum value of 1.

### ANNs

The composer paths are trained and tested on TDNN, TCN, LSTM, and Morphognosis ANNs.

### Time delay neural network (TDNN)

The TDNN, introduced in the 1980s (Waibel et al., 1989), is a multilayer architecture that classifies unsegmented temporal patterns, such as speech. This mean that a TDNN avoids having to determine the beginning and end points of a stream. In addition, past inputs form contexts within the network for classifying the current input. Figure 2 depicts a TDNN where inputs are "sliding" across the input layer over time.

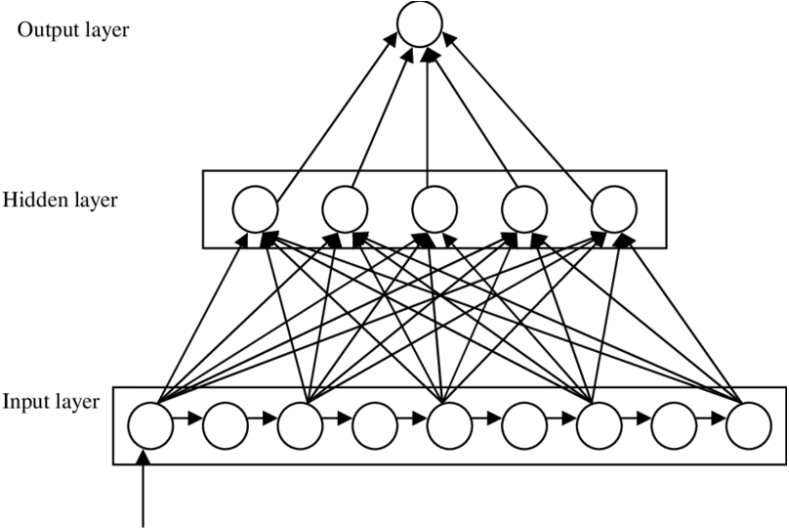

Figure 2 – TDNN network.

TEMPORAL CONVOLUTIONAL NEURAL NETWORK (TCN)

The TCN was introduced in 2018 (Bai, et al., 2018). As a convolutional network, it aims to boost performance by leveraging proximity relationships in the input. TCN is also designed as a temporal sequence classifier. It does this by the use of "causal convolutions", in which an output at a specific time can only be computed with earlier information in the stream. In addition, a TCN can span arbitrarily long input streams through the use of dilation, where higher layers receive inputs from longer spans. This shown in Figure 3, where *d* is the dilation factor.

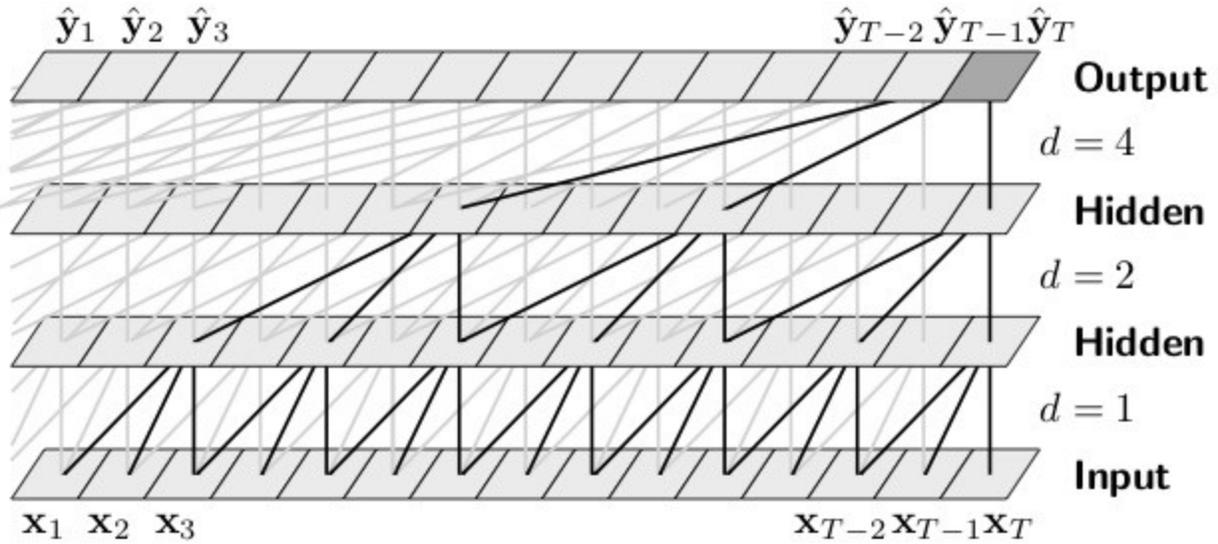

Figure 3 – TCN network.

Long short-term memory (LSTM)

The LSTM, introduced in 1997 (Hochreiter and Schmidhuber, 1997), is a recurrent neural network which has established itself as a workhorse for temporal pattern recognition. A problem in training previous recurrent neural networks is that the gradient of the error quickly vanishes as the time lag between the output and the relevant input increases, leading to the inability to train long-term state information.

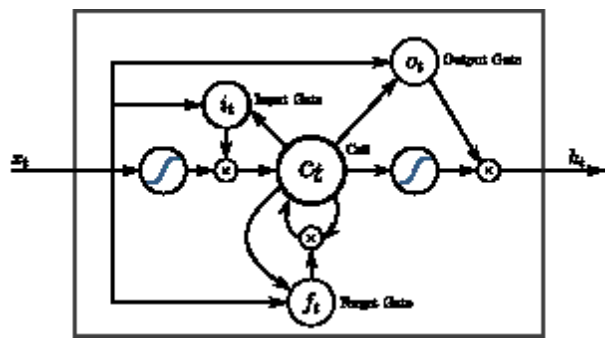

Figure 4 - LSTM memory block.

In the LSTM network, the hidden units of a neural network are replaced by memory blocks, each of which contains one or more memory cells. A memory block is shown in Figure 4. The block can latch and store state information indefinitely, allowing long-term temporal computation. What information to store, and when to output that information are part of the training process.

## MORPHOGNOSIS

Morphognosis was introduced in 2017 (Portegys, 2017). The name comes from *morphognosis* (morpho = shape and gnosis = knowledge). Morphognosis builds a spatial and temporal model of the environment that allows an organism to navigate and manipulate the environment. Its basic structure is a pyramid of event recordings called a *morphognostic*, as shown in Figure 5. At the apex of the pyramid are the most recent and nearby events. Receding from the apex are less recent and possibly more distant events.

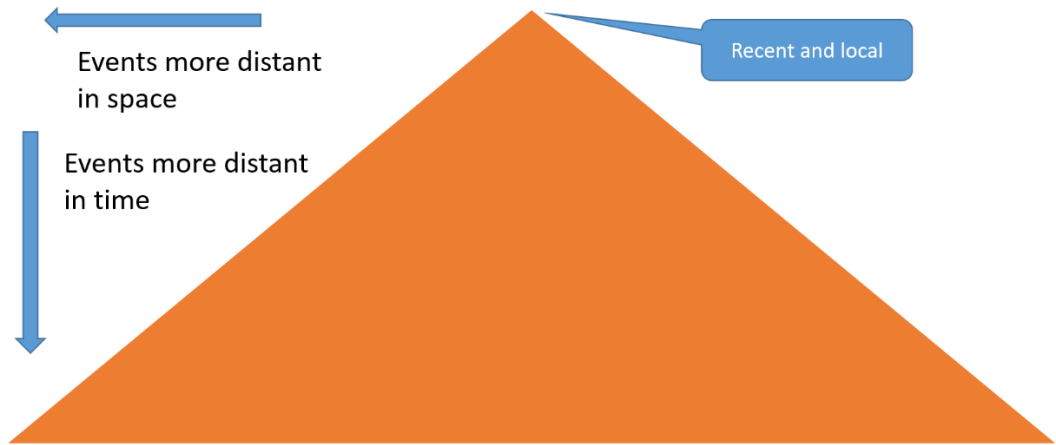

Figure 5 – Morphognostic event pyramid

Morphognosis has previously learned to guide a C. elegans nematode toward food (Portegys, 2018), build a pufferfish nest (Portegys, 2019), model a cooperative honeybee colony foraging for nectar (Portegys, 2020), and solve mazes (Portegys, 2021).

The core ANN engine for Morphognosis is a multilayer perceptron (MLP). Temporal processing is enabled in the infrastructure surrounding the MLP. This is done by building a hierarchical representation of events. In this project, the aggregation, skewing, and normalization done by the path composer accomplish this. Also, since there is no spatial component to the path task, that functionality of Morphognosis is not implemented.

## RESULTS

### ANN PARAMETERS

The ANNs were imported from the python keras 2.6.0 and tensorflow 2.6.2 packages.

The LSTM network is a sequential model, with an LSTM hidden layer of 256 neurons feeding a time distributed dense output layer. The loss is mean squared error (MSE), with the Adam adaptive momentum optimizer. It is set to issue an output for every input.

The TCN network is a sequential model with the following settings:

- kernel size = 2
- skip connection = false
- batch norm = false
- weight norm = false
- layer norm = false
- return sequences = true

The output layer is dense with linear activation.

The TDNN/Morphognosis network is a sequential model having a dense input layer with rectified linear (relu) activation feeding a hidden dense hidden layer of 256 neurons, also relu activated. The dense output layer has a sigmoid activation. The loss is binary cross entropy, and the optimizer is Adam.

### General procedure

The base path was varied at 10 and 20 steps, and the number of modular path varied at 5 and 10. The length of a modular path was randomly chosen between 2 and 5.

Training for all networks was done with 500 epochs and resulted in close to 0% errors. An error is counted when the network produces an incorrect response. 50 runs were executed to gather the results.

### Baseline performance

A baseline performance can be obtained by training the base path alone, without training the modular paths, and then testing with the modular path disruptions to the base path.

The test results are shown in Figure 6 with various mixtures of disruption types.

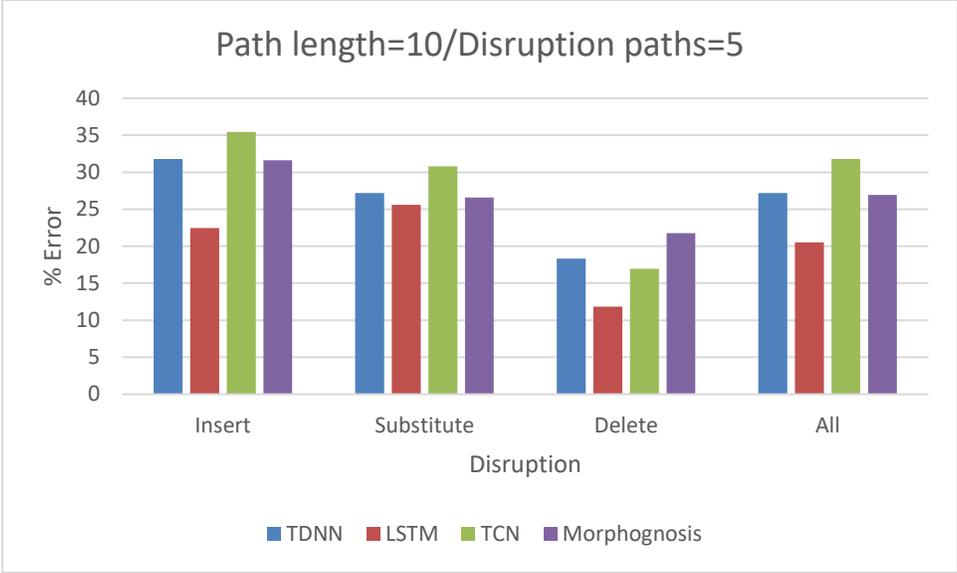

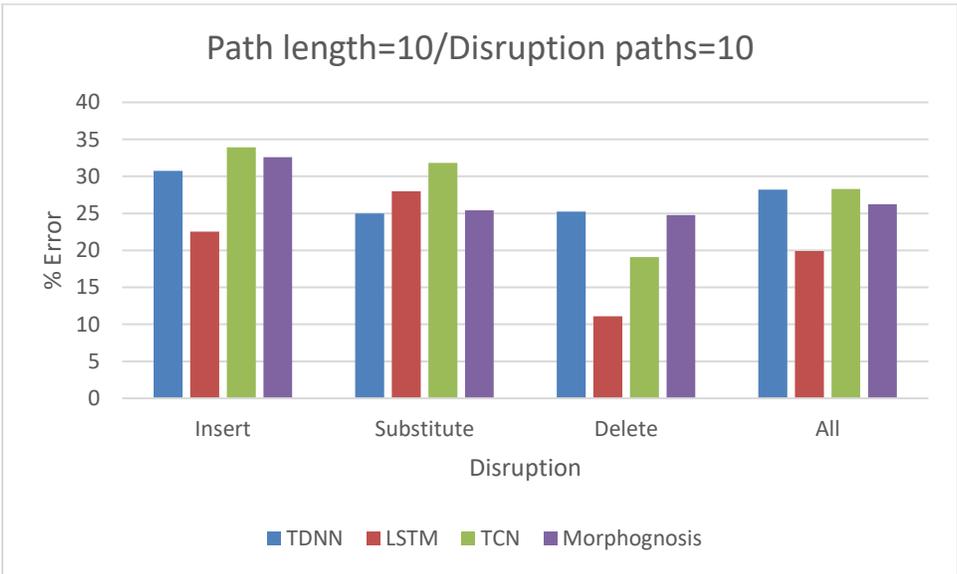

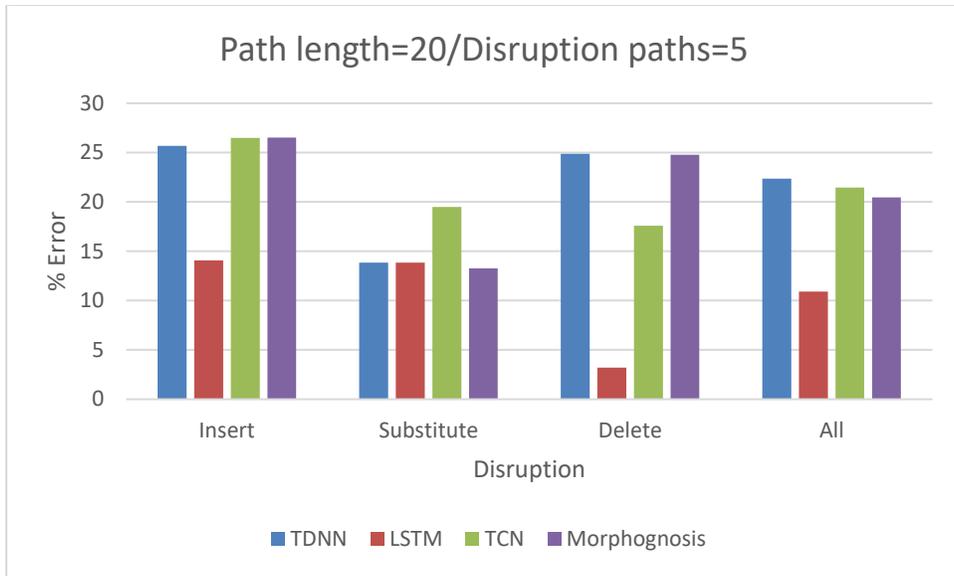

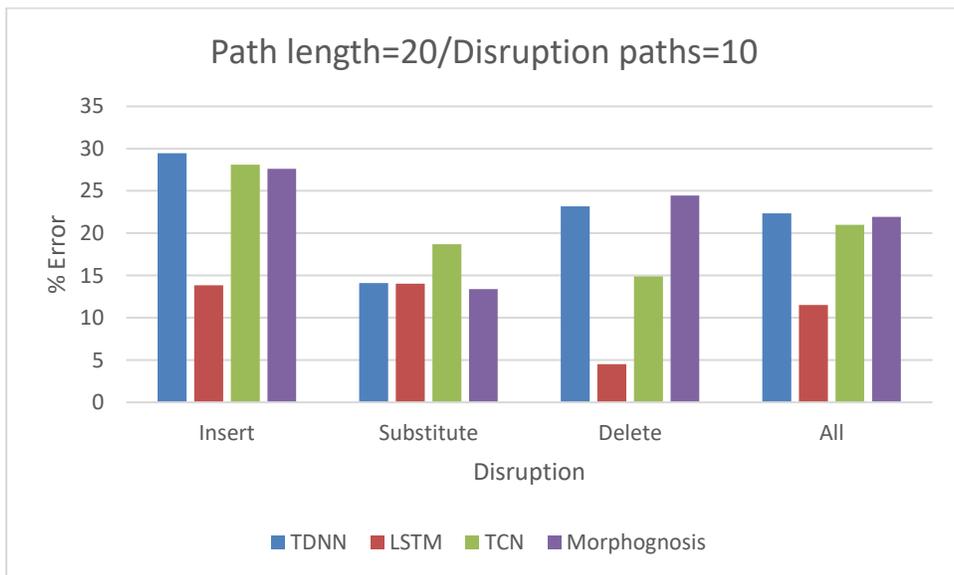

Figure 6 – Baseline testing.

## COMPREHENSIVE PERFORMANCE

The next regime tests performance with trained modular paths disrupting the base path. Figure 7 shows the testing results.

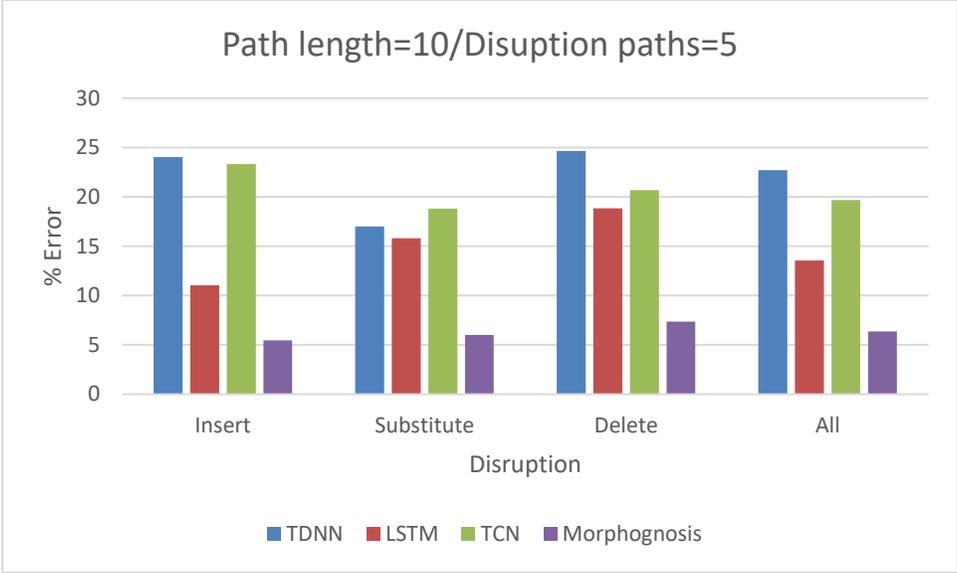
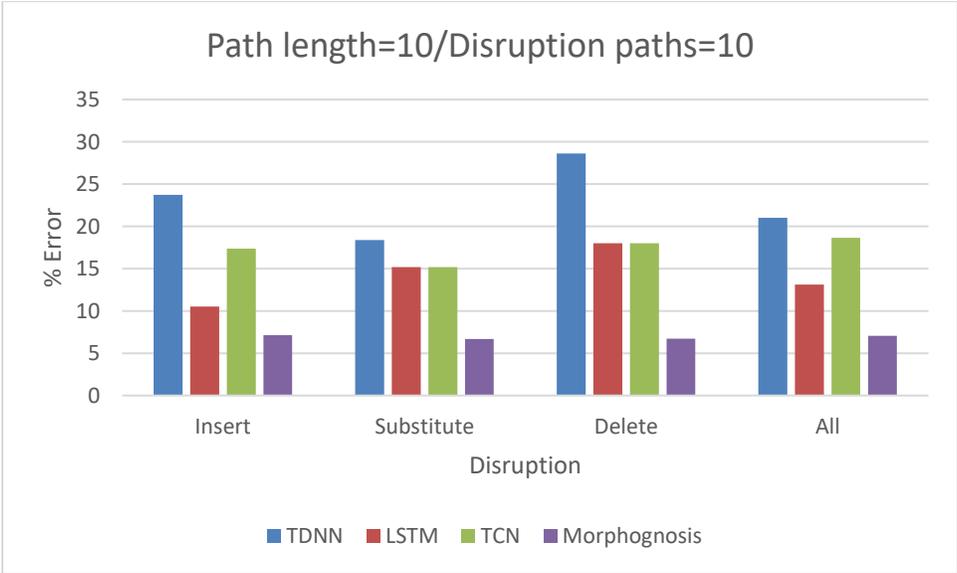

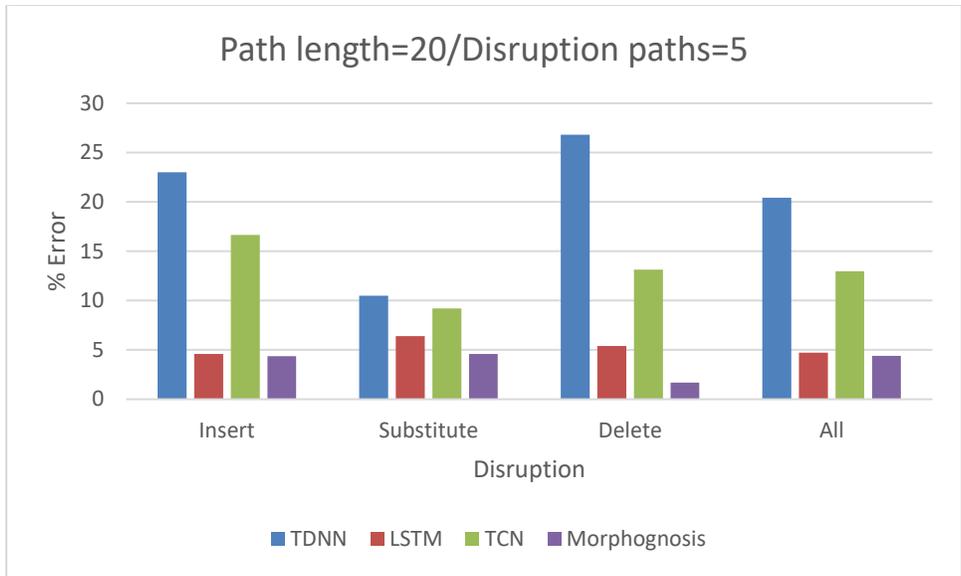

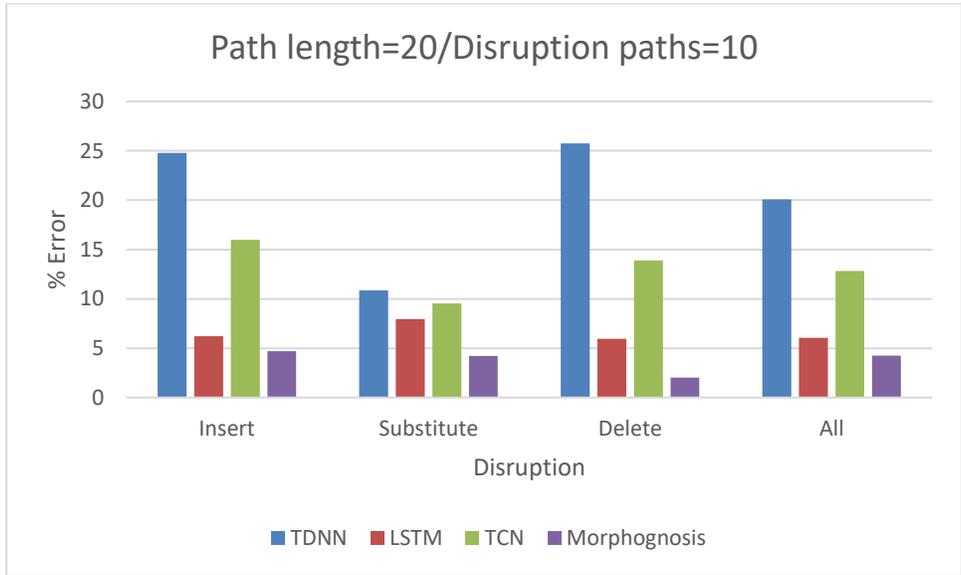

Figure 7 – Comprehensive testing.

The first thing to note is that there is a marked improvement over the baseline, unsurprisingly. It can also be seen that the LSTM and Morphognosis ANNs perform significantly better at the task. The overall ability of the networks to recover and regain tracking after disruptions is generally notable.

## FUTURE WORK

As is sometimes the case, this project "fell out" while organizing another project. That other project being a sort of capstone that mirrors a previous effort to model nesting bird behavior using the Mona goal-seeking ANN (Portegys, 2001), which is a type of modular reinforcement learning system (Moerman, 2009). The plan is to combine Mona with Morphognosis and leverage the strengths of both architectures to perform an enhanced nest-building task.

Causation is a core component of Mona, and using an MLP as an "autocoder" to efficiently identify cause-and-effect pairs seems a promising early capability of the project. This was partially inspired by Giles et al. (2001) as they explored how to use a self-organizing map (SOM) to represent financial market inputs in dimension-reduced symbols.

## CONCLUSION

For humans and many animals, their neural systems operate in a world that can be characterized as disrupted behavior streams. If our goal is to someday fully obtain this functionality for ANNs, a good start is to evaluate the current state of things. Although it is not known how well a mouse or rat would perform in the task described herein (but it would be very interesting to know), the results given here hint that ANNs are perhaps not as far from their biological counterparts as one might suppose, especially for the LSTM and Morphognosis ANNs.

## REFERENCES

Alonso, A., Bokeria, L., van der Meij, J., Samanta, A., Eichler, R., Lotfi, A. Spooner, P., Lobato, I. N., Genzel, L. (2021). The HexMaze: A Previous Knowledge Task on Map Learning for Mice. Cognition and Behavior. https://doi.org/10.1523/ENEURO.0554-20.2021

Bai, S., Kolter, J. Z., Koltun, V. (2018). An Empirical Evaluation of Generic Convolutional and Recurrent Networks for Sequence Modeling. *arXiv*. https://doi.org/10.48550/arXiv.1803.01271

Baldwin, C. Y. and Clark, K. B. (2000). *Design rules / the power of modularity*. MIT Press.

Clune, J., Mouret, J.-B., and Lipson, H. (2013). The evolutionary origins of modularity. *Proceedings of the Royal Society B: Biological Sciences*, 280(1755), Mar 2013. ISSN 1471-2954. doi:10.1098/rspb.2012.2863.

Csordás R. van Steenkiste, S., Schmidhuber, J. (2020). Are Neural Nets Modular? Inspecting Their Functionality Through Differentiable Weight Masks. *2020 ICML Workshop on Human Interpretability in Machine Learning (WHI 2020).*

Dridi, S. and Lehmann, L. (2016). Environmental complexity favors the evolution of learning. *Behavioral Ecology* 27(3), 842‑850. doi:10.1093/beheco/arv184

Giles, C.L., Lawrence, S. & Tsoi, A.C. (2001). Noisy Time Series Prediction using Recurrent Neural Networks and Grammatical Inference. *Machine Learning* **44,** 161–183. https://doi.org/10.1023/A:1010884214864

Hochreiter, S., and Schmidhuber, J. (1997). Long short-term memory. *Neural Computation*, 9(8), 1735_1780.

Hörndli, C. N. S., Wong, E., Ferris, E., Rhodes, A. N., Fletcher, T., Gregg, C. (2018). Machine Learning Reveals Modules of Economic Behavior from Foraging Mice. *bioRxiv*. doi: https://doi.org/10.1101/357434

Lorenz, D. M., Jeng, A., and Deem, M. W. (2011). The emergence of modularity in biological systems. *Physics of life reviews*, 8(2):129–160, Jun 2011.

Mearns, D. S., Semmelhack, J. L., Donovan, J. C., Baier, H. (2019) Deconstructing hunting behavior reveals a tightly coupled stimulus-response loop. *bioRxiv*. https://doi.org/10.1101/656959

Melin, P., & Castillo, O. (2005). Modular Neural Networks, 109–129. https://doi.org/10.1007/978-3-540-32378-5_6

Moerman, W. (2009). Hierarchical Reinforcement Learning: Assignment of Behaviours to Subpolicies by Self-Organization. https://www.researchgate.net/publication/242699455_Hierarchical_Reinforcement_Learning_Assignment_of_Behaviours_to_Subpolicies_by_Self-Organization

Portegys, T. E. (2001). Goal-Seeking Behavior in a Connectionist Model. *Artificial Intelligence Review.* 16 (3):225-253.

Portegys, T. (2010). A Maze Learning Comparison of Elman, Long Short-Term Memory, and Mona Neural Networks. *Neural Networks*.

Portegys, T. (2017). Morphognosis: the shape of knowledge in space and time. *The 28th Modern Artificial Intelligence and Cognitive Science Conference (MAICS)*, Fort Wayne Indiana, USA.

Portegys, T. (2018). Learning C. elegans locomotion and foraging with a hierarchical space-time cellular automaton.  *Neuroinformatics 2018 Montreal*. *F1000Research* 2018, **7**:1192 (doi: 10.7490/f1000research.1115884.1)

Portegys, T. (2019). Generating an artificial nest building pufferfish in a cellular automaton through behavior decomposition. *International Journal of Artificial Intelligence and Machine Learning (IJAIML)* 9(1) DOI: 10.4018/IJAIML.2019010101.

Portegys, T. (2020). Morphognostic honey bees communicating nectar location through dance movements. *bioRxiv*. https://doi.org/10.1101/2020.03.14.992263

Portegys, T. (2021). A modularity comparison of Long Short-Term Memory and Morphognosis neural networks. *arXiv*. https://doi.org/10.48550/arXiv.2104.11410

Rama, E., Capi, G., Mizusaki, S., Tanaka, N., Kawahara, S. (2016). Effects of Environment Changes on Rat`s Learned Behavior in an Elevated Y-Maze. *Journal of Medical and Bioengineering Vol. 5, No. 2.*

Rosenberg, M., Zhang, T., Perona, P., Meister, M. (2021). Mice in a labyrinth show rapid learning, sudden insight, and efficient exploration. *eLife*. 10:e66175. DOI: https://doi.org/10.7554/eLife.66175

Rojas, R. (1996). Neural Networks, Springer-Verlag, Berlin, Chapter 16. Modular Neural Networks.

Waibel, A., Hanazawa, T., Hinton, G., Shikano, K., Lang, K. J., (1989). Phoneme Recognition Using Time-Delay Neural Networks. *IEEE Transactions on Acoustics, Speech, and Signal Processing*, Volume 37, No. 3, pp. 328. - 339.